# A Theory on AI Uncertainty Investigation Based on Rademacher Complexity and Shannon Entropy

**Mingyong Zhou**

School of Computer Science and Communication Engineering
GXUST
Liuzhou, China
Zed6641@hotmail.com

**Abstract.**

In this paper, we present a theoretical discussion on AI deep learning neural network uncertainty investigation based on the classical Rademacher complexity and Shannon entropy. First it is shown that the classical Rademacher complexity and Shannon entropy is closely related by quantity by definitions. Secondly based on the Shannon mathematical theory on communication [3], we derive a criteria to ensure AI correctness and accuracy in classifications problems. Last but not the least based on Peter Barlette's work in [1], we show both a relaxing condition and a stricter condition to guarantee the correctness and accuracy in AI classification . By elucidating in this paper criteria condition in terms of Shannon entropy based on Shannon theory, it becomes easier to explore other criteria in terms of other complexity measurements such as Vapnik-Cheronenkis, Gaussian complexity by taking advantage of the relations studies results in other references such as in [2] . A close to 1/2 criteria on Shannon entropy is derived in this paper for the theoretical investigation of AI accuracy and correctness for classification problems.

# 1 Introduction

AI uncertainty investigation becomes more and more important while AI deep learning neural networks are applied successfully in more and more areas including imaging diagnosis and pattern classification problems etc. Though AI achieves great success in applications covering huge industries and applications, the theoretical investigation for AI success is less performed as compared to the applications investigations. However the theory study on how and why AI is so successful is by all means very important in order to understand the mechanism of success of the AI deep learning neural networks, so that in future AI can be understood better and applied better with theoretical guidance for industry applications.

In this paper based on Shannon entropy and Shannon theory [3], we provide a different perspective and derive a criteria for AI uncertainty in classification problems. From our perspective, we are able to build a theory on AI uncertainty based on Shannon mathematical theory on communications. First classical Rademacher complexity and Shannon entropy are investigated for their quantity by definitions. It is shown that they are closely related with each other. Secondly based on the Shannon mathematical theory on communication, we are able to derive a criteria to ensure AI correctness and accuracy in classifications problems by comparing an AI to a communication channel and define new AI quantities such as in communication channels. Last but not the least based on Peter Barlette's work in [1], we show both relaxing conditions and much more strict conditions to guarantee the correctness and accuracy in AI classifications. By elucidating the criteria condition in terms of Shannon entropy based on Shannon theory, it becomes easier to explore other criteria in terms of other complexity measurements such as Vapnik-Cheronenkis, Gaussian complexity by taking advantage of the relations studies results in other references . In this paper a 1/2 criteria of Shannon entropy is derived for the theoretical investigation of AI accuracy and correctness for classification problems.

In **Section 2** , staring with the Shannon definitions on channel capacity and rate and its main result, we derive a basic criteria for misclassification by observing a quantity relation between Rademacher complexity and Shannon entropy .In **sections 3**, a discussion is devoted to the error upper and lower bound ranges effects on the criteria based on reference [1] . Both a relaxing condition and a much more strict criteria condition based on the AI error upper bound and lower bounds are derived. In **section 4** future work direction is indicated. This paper is concluded with **section 5** in which major results are summarized.

## 2  A Basic Criteria for AI Misclassifications

In this section, we first overview the main results of Shannon theory by reviewing Shannon's definition on channel capacity and information rate. Then we present an AI model for pattern classification problems and show how the AI model can be compared to a communication channel in terms of Shannon entropy. We will show that AI model and communication channel are closely correlated in principle. Further we elucidate a quantity relation between classical Rademacher complexity and Shannon entropy from their definitions before we derive a basic close to 1/2 criteria for AI misclassification.

$$R < C$$
$$C = Max(H(x) - H_y(x)) \qquad (2.1)$$

where $C$ is defined as the channel capacity in Shannon theory for a noisy channel, and $R$ is channel transmission rate. Equation (2.1) by Shannon theory indicates that the transmission rate $R$ must not exceed channel capacity $C$ in order to recover the original information by any desired accuracy [3]. Now with an AI mode for classification problems, one is faced with problems as follows: given an AI deep learning neural network with the following error ranges

$$H(e_{min}) < H_y(x) < H(e_{max}) \qquad (2.2)$$

and a complex classification probelm with certain Rademacher complexity,one is required to find a criteria by which one can classify the patterns correctly with any accuracy desired. Obviously $R$ rate in Shannon theory is closely related to the Rademacher complexity for a given complex problem. To show this observation, we only need to notice the defintion of classical Rademacher complexity $R(F)$ by equation (2.3), that is, the quantity $R(F)$ reaches the maximum of 1.0 in case that the most complex binary +1, -1 are matched exactly. This is equivalent to the communication channel by which the information source is transmitted to reach the receiver with no errors. Thus there is a very close relations between Shannon's rate and classical Rademacher complexity by quantity. The higher the rate $R$ in communication, the more complex $R(F)$ in AI model for pattern classification problems, where both quantity reaches 1.0 as mximum values. At the same time, we also notice that there is also quantitive relationship between $R(F)$ and $H_y(x)$, by definitons of Shannon entropy $H_y(x)$ reaches maximum value of 1.0 indicating the most "chaos" scenario, and a minimum value of 0 otherwise .Therefore we conclude that three quantities R rate in Shannon theory, $R(F)$ as Rademacher complexity for classification problems and $H_y(x)$ are closely related in quantity. In fact the three quantities are almost equivalent to each other as demonstrated in Figure 1 , Figure 2 and Figure 3.

$$R(F) = E\{\tfrac{1}{n}\sup_{f \in F} [\sigma_i \times f(X_i)]\}, \sigma_i = \pm 1, f(X_i) = \pm 1$$

(2.3)

With the above observations on three quantities, we now derive a basic criteria for AI misclassifications. By Shannon theory on the noise channel with conditional entropy $H_y(x)$, to ensure transmit with no errors starting from equation (2.1), the following (2.4) must be satisfied

$$R < Max(H(x)) - \min H_y(x)$$

(2.4)

Equation (2.4) implies that

$$R + \min H_y(x) < Max(H(x))$$

(2.5)

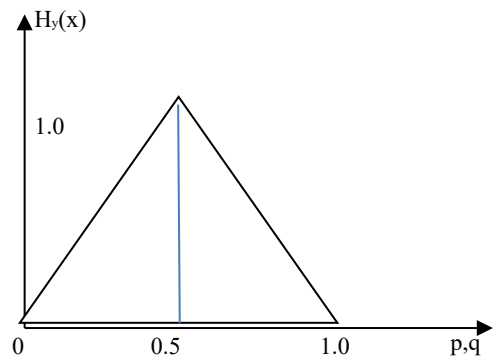

Figure 1: Shannon Conditional entropy Hy(x)

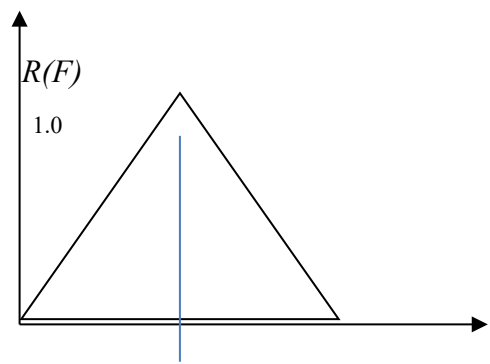

0  f(xi)

Figure 2: Rademacher Complexity R(F)

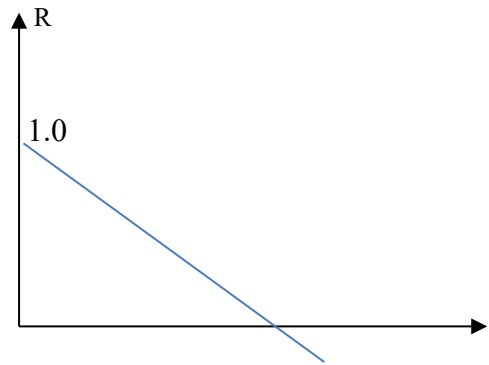

Figure 3: Shannon's Channel Rate R

With the equivalence between R in Shannon theory on communication and *R(F)* of classical Rademacher complexity, we have

$$R(F) + \min H_y(x) < Max(H(x))$$

(2.6)

The analysis is performed as part A) and part B):

A) , if  R(f)> min (Hy(x)), then

$$2\min(H_y(x)) \leq \min(H_y(x)) + R(f) \leq Max(H(x))$$
$$\min(H_y(x)) \leq \frac{1}{2} Max(H(x))$$

(2.7)

B), if  R(f)<min (Hy(x)),  then

$$R(f) \leq \min(H_y(x)) \leq Max(H(x)) - R(f)$$
$$R(f) \leq \frac{1}{2} Max(H(x))$$

(2.8)

Therefore either equation (2.7) or (2.8)  must  be satisfied . For simplicity

$$\min(H_y(x)) \leq \frac{1}{2}, (Max(H(x) = 1.0)$$

(2.9)

Or from part B) analysis

$$R(f) \leq \frac{1}{2}, (Max(H(x) = 1.0))$$

(2.10)

The analysis from equations (2.9) and (2.10) implies that given *R(f)* as the Rademacher complexity of a classification problem and an upper and lower bounds denoted by equation (2.2), the lower bound in equation (2.2) regarding $H_y(x)$ and *R(f)* must satisfy either equation(2.9) or (2.10) to ensure an AI classification with any accuracy desired. We term it as 1/2 criteria in our paper. It should be noted that these are very relaxing conditions that only require either lower bound of error to satisfy an upper bound or Rademacher complexity of given problem *R(f)*<1/2. In real applications, however, we handle complex classification problems. Therefore we can assume that *R(f)*>1/2 in real applications. In fact one can as well impose a much more strict condition on part A) by the following

$$H(e_{max}) < \tfrac{1}{2}, (Max(H(x)) = 1.0)$$

(2.11)

Equation (2.11) implies that Shannon entropy of all AI errors lies below the value of 1/2 while equation (2.9) indicates that only lower bound of AI errors and its Shannon entropy lies below the value of 1/2. Thus equation (2.11) is thus a much more strict condition for no AI mistakes in pattern classifications.

# 3 A Further Discussion on the Criteria Based on Error's Upper and Lower Bound for AI Model

According to Peter Barlette's analysis[1] on the error bounds for neural networks,

an upper bound and lower bound is perspectivelly derived as follows.

$$e_{\min} = e_2 + \max(\tfrac{d-1}{32m}, \tfrac{7\ln(1/\delta)}{8m})$$

(3.1)

$$e_{\max} = e_2 + \sqrt{\tfrac{c}{m}(\tfrac{A^2 n}{r^2}\log(\tfrac{A}{r})\log^2 m + \log(1/\delta))}$$

(3.2)

By equation (2.9)

$$H(e_{\min}) = H(e_2 + \max(\tfrac{d-1}{32m}, \tfrac{7\ln(1/\delta)}{8m})) < \tfrac{1}{2}Max(H(x))$$

(3.3)

where m is the training different patterns number, A is defined as size of neural network weight size indicated in [1]. e2 's are the training squared error for all training samples. A more strict condition is as follows.

$$H(e_{max}) = H(e_2 + \sqrt{\frac{c}{m}(\frac{A^2n}{r^2}\log(\frac{A}{r})\log^2 m + \log(1/\delta))}) < \frac{1}{2}Max(H(x))$$

(3.4)

At the first glance, the result derived in this paper is more or less puzzling in that a lower bound of error whose Shannon entropy has an upper bound is enough for a guaranteed accuracy for AI classifications with any accuracy dersired. But if one thinks about the real applications, it does make sense in that if an AI reaches about Shannon entropy 1/2, the probability of error is less than about 10% for 1 and 0 two patterns. That means an AI can almost recognize the correct patterns with no mistakes. In short we can impose a more strict condition as follows to ensure all AI model error's Shannon entropy is less than 1/2.

$$H(e_2 + \sqrt{\frac{c}{m}(\frac{A^2n}{r^2}\log(\frac{A}{r})\log^2 m + \log(1/\delta))}) < \frac{1}{2}$$
$$(MaxH(x) = 1)$$

(3.5)

According to the complex analysis in [1], m's are the number of distinctive different patterns should be kept large enough so that the upper bound of AI model errors can be kept small. The size of weights indicated as A in equation (3.5) should be kept as small as well in trainings so that the overall error's upper bound can be kept as small as possible.

## 4 Future work

Based on the derived results in this paper, we can extend to other complexity measures such as Vapnik-Cheronenkis(VC) and Gaussian complexity by provoking the results in [2]. Criteria based on other measures are possible if one look into the quantity relationships such as in [2].

## 5 Conclusions

We show basic quantity relationships between classical Rademacher complexity and Shannon entropy. First classical Rademacher complexity is outlined and its quantity relations with Shannon entropy is elucidated. Secondly by comparing AI to a communication channel, we are able to apply Shannon theory and entropy to investigate the AI uncertainty. A 1/2 criteria is derived from Shannon theory on noise communication channels, that is , either Shannon entropy of AI model error lower bound is less than 1/2, or the Rademacher complexity of given problem is less than 1/2. The theoretical results in this paper can provide a guidance for AI applications and model selections to ensure no ambiguity in AI outcomes. The main results are summarized as follows:

(1), Classical Rademacher complexity is almost equivalent to Shannon entropy in quantity. In fact three quantities Rademacher complexity, Shannon conditional entropy $H_y(x)$ and rate R are closely correlated in quantity.

(2), If AI capability C is defined as in Shannon theory, either $\max(H_y(x)) < 1/2$ or $R(f) < 1/2$ can ensure that AI model can restore original information with any small errors desired, (In fact if $R(f) > 1/2$ is observed in real applications, only $\max(H_y(x)) < 1/2$ is required for this purpose which is a much more strict condition)

12(3), A pragmatic approach is proposed for AI deep learning neural network, that is, assumed under the condition of R(f) >1/2 for real application problems by training well (in the sense e.g. to keep the size of weights small enough) the AI by large distinctive samples [1] and by testing the error probability to ensure $\max(H_y(x)) < 1/2$ so that in applications AI can attain the most desired performance with any small errors desired according to Shannon Theory.

It should be emphasized that the results derived in this paper including 1/2 criteria are built upon Shannon mathematical theory on communication.